\pdfoutput=1 
\documentclass[a4paper,twoside]{article}
\usepackage{epsfig}
\usepackage{subcaption}
\usepackage{calc}
\usepackage{amssymb}
\usepackage{amstext}
\usepackage{amsmath}
\usepackage{amsthm}
\usepackage{multicol}
\usepackage{pslatex}
\usepackage{apalike}
\usepackage{setspace}
\usepackage{xcolor}
\usepackage[bottom]{footmisc}
\usepackage{SCITEPRESS}     % Please add other packages that you may need BEFORE the SCITEPRESS.sty package.

\begin{document}

\title{Estimation of Looming from LiDAR}

\author{\authorname{\textcolor{black}{Juan D. Yepes and Daniel Raviv}}
\affiliation{\textcolor{black}{Florida Atlantic University \\
	Electrical Engineering and Computer Science Department\\ 
Boca Raton, FL, USA}}
\email{\textcolor{black}{jyepes@fau.edu, ravivd@fau.edu}}
}


%% ABSTRACT
\abstract{Looming, traditionally defined as the relative expansion of objects in the observer's retina, is a fundamental visual cue for perception of threat and can be used to accomplish collision free navigation. The measurement of the looming cue is not only limited to vision and can also be obtained from range sensors like LiDAR (Light Detection and Ranging). In this article we present two methods that process raw LiDAR data to estimate the looming cue. Using looming values, we show how to obtain threat zones for collision avoidance tasks. The methods are general enough to be suitable for any six-degree-of-freedom motion and can be implemented in real-time without the need for fine matching, point-cloud registration, object classification or object segmentation. Quantitative results using the KITTI dataset shows advantages and limitations of the methods.}

\onecolumn \maketitle \normalsize \setcounter{footnote}{0} \vfill

%% INTRODUCTION
%\onecolumn
\section{\uppercase{Introduction}}
\label{sec:introduction}

%STEP1 - Hooks
% UNRESOLVED ISSUE
Collision avoidance continues to be one of the greatest challenges in unmanned vehicles \cite{yasin2020unmanned} and autonomous driving \cite{roriz2021automotive}, where the demand for increasingly robust, fast and safe systems is crucial for the development of these industries. It is closely tied with the perception of threat. For creatures in nature, this task seems to be more fundamental than the recognition of shapes \cite{albus1990motion}. Studies in biology have shown strong evidence of neural circuits in the brains of creatures related to the identification of looming \cite{ache2019neural}. Basically, creatures have evolved instinctive escaping behaviors that tie perception directly to action. In this way they can avoid imminent threat from predators that project an expanding image on their visual systems \cite{evans2018synaptic} \cite{yilmaz2013rapid}.

% STEP3 - POSITION YOUR APPROACH 
% More background summary on the Looming cue
The visual looming cue, defined quantitatively by \cite{raviv1992quantitative} as the instantaneous change of range over the range, is related to the increase in size of an object projected on the observer's retina. It is an indication of threat that can be used to accomplish collision avoidance tasks.

% Introduce LiDAR
The looming cue is not limited exclusively to vision and can be computed from active sensors like LiDAR (Light Detection and Ranging). LiDAR-systems are very popular in autonomous vehicles since they are less sensitive to lighting conditions, have wider field of view (FOV) than cameras and require less processing time and computation power \cite{wang2018robust}. 

% Introduce Layers in autonomous systems, and the perception Layer
In general, architectures for autonomous vehicles consist of three sub-systems: Perception, Planning and Control \cite{sharma2021recent}. The approaches for implementing the Perception sub-system can be mainly divided subsequently into three categories: camera-based, LiDAR-based and sensor fusion-based approaches \cite{wang2018robust}.

% Introduce the main aproaches for collision avoidance
Traditionally the task of obstacle avoidance is exclusively related to the Planning and Control sub-systems where the action can be categorized further into four major approaches: Geometric, Force-Field, Optimized, Sense and Avoid \cite{yasin2020unmanned}. The Planning and Control sub-systems demand from the Perception sub-system, as a prerequisite, the detection and classification of objects in 3D or 2D space prior to any global or local path planning action \cite{mujumdar2011evolving}.

In contrast, it has been shown that obstacle avoidance tasks can be accomplished without 3D reconstruction (structure from motion), using the Active Perception paradigm. The idea is to estimate normal flow from derivatives of image intensity and from there an estimation of time-to-contact with features in the scene allows for local navigation actions that steer the robot away from the hazard \cite{aloimonos1992visual}.

% STEP 4 - SPECIFIC RESEARCH PROBLEM
In this paper we follow an approach similar to \cite{Raviv2000TheVL}. It uses 2D raw data to directly get quantitative indication of visual threat. We present methods for obtaining looming values using raw LiDAR data and threat zones. This allows for real-time collision avoidance tasks without 3D reconstruction or scene understanding. In the proposed approach further processing of the LiDAR point cloud is not required for obtaining threat zones.

%% RELATED WORK
\section{\uppercase{Related Work}}

% FIGURE EXPANDING TREE ON SPHERE
\begin{figure}
	\centering
	{\epsfig{file = 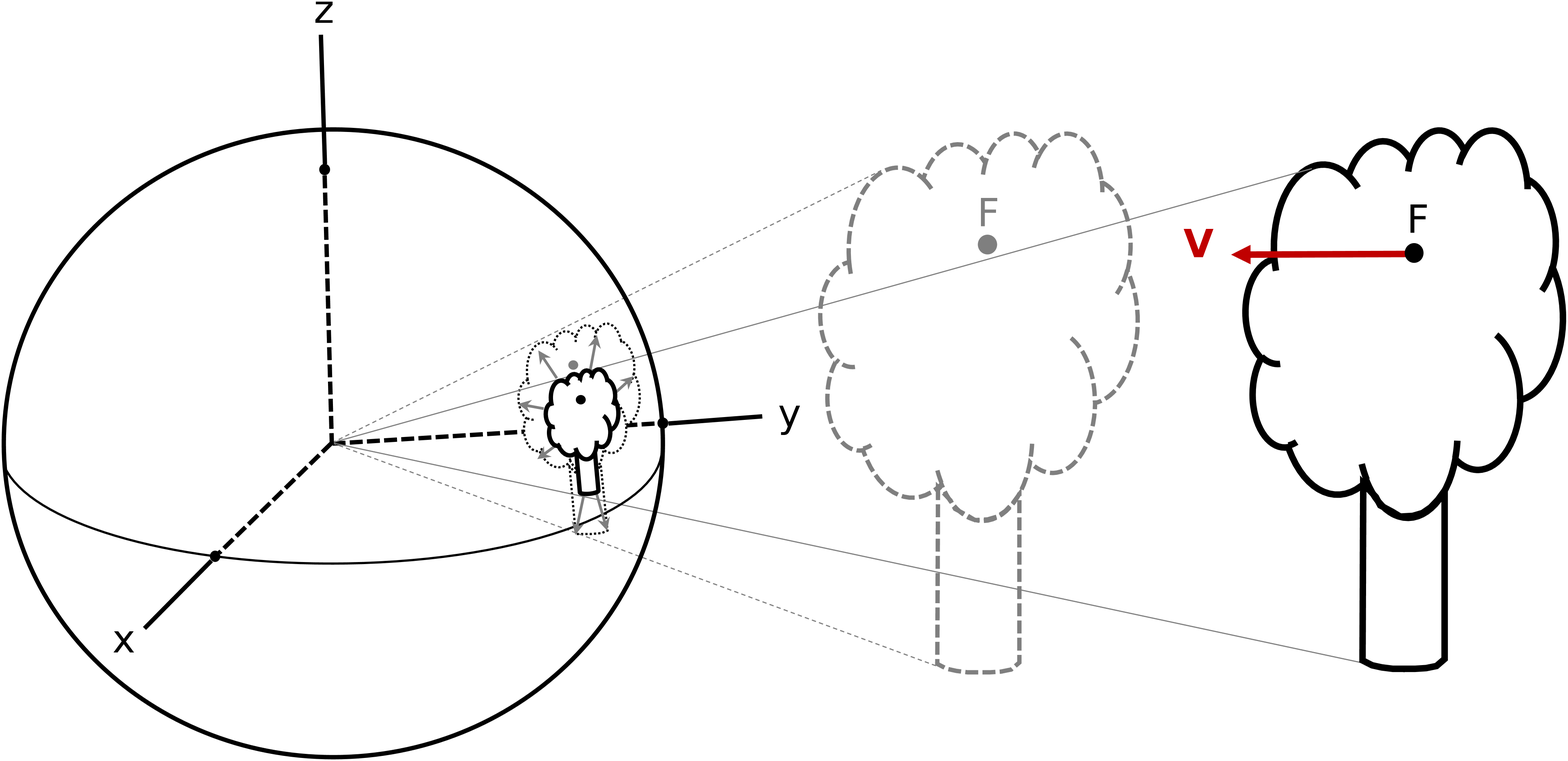, width = 78mm}}
	\caption{Visualizing looming cue: relative expansion of 3D object on 2D image sensor.}
	\label{Fig:LoomingCue}
\end{figure}

\subsection{Visual looming}
%previuos methods to estimate looming\\
The visual looming cue is related to the relative change in size of an object projected on the observer's retina as the range to the object varies (Figure \ref{Fig:LoomingCue}). It is \textit{quantitatively} defined as the negative value of the time derivative of the range between the observer and a point in 3D divided by the range \cite{raviv1992quantitative}:
% EQUATION LOOMING FROM LIMITS
\begin{align}
	L &= - \lim\limits_{\Delta t \to 0}\frac{\left(\frac{r_2-r_1}{\Delta t}\right)}{r_1}  \label{E:LoomingDiscrete} 
\end{align}

where $L$ denotes looming, $t_1$ represents time instance 1, $t_2$ represents time instance 2, $\Delta t$ is $t_2 - t_1$, $r_1$ is the range to the point at time instance $t_1$, and $r_2$ is the range at time instance $t_2$.\\

In the limit $L$ can also be expressed as:   					

% EQUATION RDOT OVER R
\begin{align}
	L &= -\left(\frac{\dot{r}}{r}\right)  \label{E:Looming} 
\end{align}

where dot denotes derivative with respect to time. The reason for the negative sign in equations \eqref{E:LoomingDiscrete} and \eqref{E:Looming} is to associate image expansion with positive values of looming.

The looming $L$ can also be expressed in vector form as:
\begin{align}
	L &= \frac{\mathbf{t}\cdot \mathbf{r} }{\mathbf{r}\cdot \mathbf{r}}  \label{E:LoomingVector} 
\end{align}
\par where $\mathbf{t}$ is the instantaneous translation velocity vector of the vehicle and $\mathbf{r}$ is the relative position vector of the point with respect to the vehicle origin.

% PROPERTIES OF LOOMING
\subsubsection{Looming properties}

Note that the result for $L$ in equation \eqref{E:LoomingVector} is a scalar value which is \textit{dependent} on the vehicle translation component but \textit{independent} of the vehicle rotation.
Also, $L$ is measured in $[time^{-1}]$ units.

It was shown that points in space around the vehicle that share the same looming values form equal looming spheres with centers that lie on the instantaneous translation vector $\mathbf{t}$ and intersect with the vehicle origin. These looming spheres expand and contract depending on the magnitude of the translation vector \cite{raviv1992quantitative}.

\begin{figure}
	\centering
	{\epsfig{file = 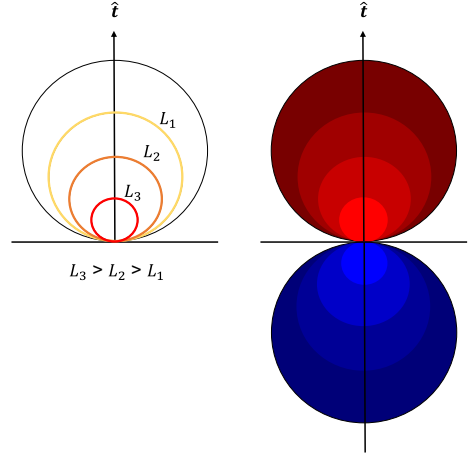, width = 60mm}}
	\caption{Equal looming spheres as shown in red for $(L>0)$ and blue for $(L<0)$.}
	\label{fig:EqualLoomingCircles}
\end{figure}

Since an equal looming sphere corresponds to a particular looming value, there are other spheres with varying values of looming with different radii. A smaller sphere signifies a higher value of looming as shown in Figure \ref{fig:EqualLoomingCircles} ($L_3 > L_2 > L_1$).

Regions for obstacle avoidance and other behavior related tasks can be defined using equal looming surfaces. For example, a high danger zone for $L>L_3$, medium threat for $L_3 > L > L_2$ and low threat for $L_2 > L > L_1$.

% FIGURE EQUAL LOOMING CIRCLES AND COLORS

\subsubsection{Advantages of looming}
Sensory data from LiDAR systems provide point clouds that can be used for 3D reconstruction. Using information about ego-motion combined with 3D reconstruction it can help in scene understanding (using mainly machine learning and AI techniques) followed by path planning to achieve obstacle avoidance. Note that additional processing is needed when dealing with moving objects.

Using the same LiDAR point cloud, it is possible to get looming directly without 3D reconstruction and without knowing the ego-motion of the LiDAR sensor. Looming is measured in \textit{time} units and provide information about imminent threat to the observer. There is no need for scene understanding such as identifying cars, bikes, or pedestrians. In addition, looming provides threats for moving objects as well. Extracting looming for obstacle avoidance using point cloud raw data from LiDAR is the main contribution of this paper.

\subsubsection{Measuring visual looming}
Several methods were shown to quantitatively extract the visual looming cue on a 2D image sequence by measuring attributes like area, brightness, texture density and image blur \cite{Raviv2000TheVL}.

\textit{Relative change in image area:}
Visual looming is calculated from the relative temporal change in image area and the surface orientation by assuming that the camera can be fixated on a planar patch of the object. 

\textit{Relative change in image brightness.}:
Visual looming can also be calculated from the relative rate of change of image brightness by assuming Lambertian surfaces, i.e., surfaces that are equally bright irrespective of the angle of view.

\textit{Relative change in texture density}:
Visual looming can be calculated from the texture density and its temporal changes by assuming a small area around the point of fixation to be locally planar (and the density of texture primitives on the 3D surface to be constant in that region). 

\textit{Relative change in image blur}:
The relative change in image blur or defocus is used to calculate the visual looming. It is assumed that the camera's depth of field is very small. Looming can be measured by studying the relative change in the radius of the blur circle. 

% Similar to Looming - VTC
The Visual Threat Cue (VTC), just like the visual looming, is a measurable time-based scalar value that provides some measure for a relative change in range between a 3D surface and a moving observer. The VTC is computed from relative variations of a global image dissimilarity measure called the image quality measure (IQM) that is extracted directly from gray level images. The method has been shown to be robust allowing for closed loop control of vehicles \cite{kundur1999novel}.

% Looming with event cameras
Event-based cameras were shown to detect looming objects in real-time from optical flow using clustering of pixels. The method is fast, taking only microseconds to process each event assuming that a single object is moving in the scene \cite{ridwan2018looming}.

%PREVIOUS METHODS RELATED TO LiDAR
\subsection{LiDAR for obstacle detection and avoidance}
% What is the flow from perception to obstacle avoidance using LiDAR
In general, collision avoidance approaches require 3D reconstruction of the environment where collision free paths can be computed an executed.

Several deep learning and path planning techniques for autonomous driving were mentioned in \cite{grigorescu2020survey}. The path planning process considers all possible obstacles in the environment in order to calculate a trajectory along a collision‐free route.

General approaches for collision avoidance that use reactive, deliberate or hybrid planning where identification of objects is a requirement before planning an evasive action were mentioned in \cite{yasin2020unmanned}. 

An online 2D LiDAR-based object detection and tracking approach using cylinders to model obstacles in the environment is proposed in \cite{wang2018robust}. Point clouds, provided by LiDAR sensors, are the preferred representation for many scene understanding related applications. These point clouds are processed to build real-time 3D localization maps, using SLAM (Simultaneous Localization and Mapping) and related techniques. A LiDAR-only SLAM method called LOAM that estimate vehicle velocity by odometry and map the environment in real-time using fine matching and point cloud registration was developed by \cite{zhang2014loam}. 

Several deep learning methods for 3D understanding, including 3D shape classification, 3D object detection and tracking were presented by \cite{guo2020deep}. Methods for ground segmentation and identification of drive-able space were mentioned by \cite{roriz2021automotive}.

In general, most methods use 3D object detection, segmentation and scene understanding. In addition, ego-motion is required for obstacle avoidance and autonomous navigation.

In contrast, our approach provides relevant information for the task of collision avoidance in the form of a real-time looming threat map. This allows the direct transition from perception to action without the need of object identification, finding ego-motion or 3D reconstruction.

%% METHOD - LOOMING FROM LIDAR
\section{\uppercase{Method}}

\subsection{Derivation of looming from the relative velocity field}
We will derive a general expression for Looming ($L$) for any six-degree-of-freedom motion that involves velocity and range using spherical coordinates. Then we can apply this expression with LiDAR data.

% VEHICLE MOTION
\paragraph{Vehicle motion} Consider a vehicle undergoing a general six-degree-of-freedom motion in a 3-D space relative to an arbitrary stationary reference point. 

% FIGURE TRAJECTORY OF A VEHICLE
\begin{figure}
	\centering
	{\epsfig{file = 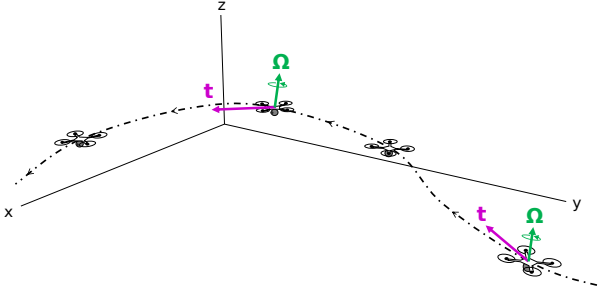, width = 78mm}}
	\caption{Vehicle undergoing a generalized six-degree-of-freedom motion relative to the \textit{world frame}. $\mathbf{t}$ is the translation velocity vector, and ${\Omega}$ is the rotation vector.}
	\label{trajectory}
\end{figure}

At any given time, the vehicle will have an associated translation velocity vector $\mathbf{t}$ and a rotation vector ${\Omega}$ relative to the \textit{world frame} as shown on Figure \ref{trajectory}.\\

% FIGURE 3D AXIS
\begin{figure}
	\centering
	{\epsfig{file = 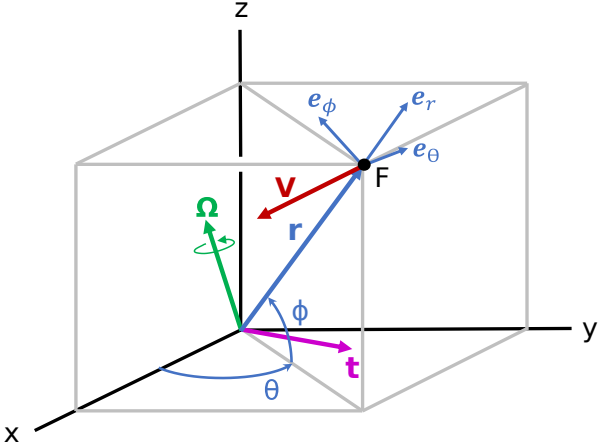, width = 78mm}}
	\caption{LiDAR coordinate system.}
	\label{CoordinateSystem}
\end{figure}

\paragraph{LiDAR coordinate system}
Consider a local coordinate system centered at the LiDAR sensor, fixed to the moving vehicle. We chose the x-axis to be aligned with the forward direction of the vehicle and the z-axis to be aligned with the vertical orientation.
 
Refer to Figure \ref{CoordinateSystem}: In this frame any stationary feature on the 3-D scene can be represented by spherical coordinates $(r,\theta,\phi)$ where $r$ is the radial range to the feature point $\mathbf{F}$, $\theta$ is the azimuth angle measured from the x-axis and $\phi$ is the elevation angle from the XY-plane. 
Rectilinear and spherical coordinates conversions are given by:

% RECTILINEAR COORDINATES
\begin{gather}\label{rectSpherical}
	\begin{align}
		x &= r \cos\phi\cos\theta, &r &= \sqrt{x^2 + y^2 + z^2} \\
		y &= r \cos\phi\sin\theta, &\theta &= arctan(y,x)\notag\\
		z &= r \sin\phi, &\phi &= arctan(z,\sqrt{x^2 + y^2})\notag
	\end{align}
\end{gather}
Note that in order to eliminate quadrant sign confusion we take advantage of two argument $arctan$ function for the computation of $\theta$ and $\phi$.\\

\paragraph{Relative velocity field} In our analysis the vehicle is undergoing a generalized six-degree-of-freedom motion in the\textit{ world frame}. This is equivalent to interpreting the motion as the vehicle being stationary and the feature point $\mathbf{F}$ moving on the \textit{LiDAR frame} with opposite velocity vector $\mathbf{-t}$ and rotation  ${-\Omega}$.

In this \textit{LiDAR frame} the feature $\mathbf{F}$ has relative velocity vector:

% VELOCITY EQUATION
\begin{align}
	\mathbf{V} &= (-\mathbf{t}) + (-{\Omega} \times \mathbf{r}) \label{E:VelocityField} 
\end{align}

Notice that for a given feature $\mathbf{F}$ the relative velocity vector $\mathbf{V}$ can be interpreted as the \textit{relative velocity field} in 3-D due to ego-motion. ($\mathbf{r}$ in bold refers to the range vector and the scalar $r$ to its magnitude, i.e $r = |\mathbf{r}|$).\\

We can conveniently express $\mathbf{V}$ using rectilinear or spherical unit vector components. For this purpose we can apply the derivative with respect to time of $x,y,z$ in equation set \eqref{rectSpherical}, shown here in matrix notation:

% SPHERICAL TO RECTILINEAR VELOCITY
\begin{gather}\label{sphericalToRectDot}
	\begin{bmatrix}
		\dot{x}\\
		\dot{y}\\
		\dot{z}
	\end{bmatrix}
	=
	\begin{bmatrix}
		\cos\theta\cos\phi & -\sin\theta & -\cos\theta\sin\phi\\
		\sin\theta\cos\phi & \cos\theta & -\sin\theta\sin\phi\\
		\sin\phi & 0 & \cos\phi	
	\end{bmatrix}
	\begin{bmatrix}
		\dot{r}\\
		r\dot{\theta}\cos\phi\\
		r\dot{\phi}
	\end{bmatrix}
\end{gather}

Since the inverse of the transformation matrix above is also its transpose we can easily find the correspondent components in spherical coordinates:

% RECTILINEAR TO SPHERICAL VELOCITY
\begin{gather}\label{dotComponents}
	\begin{bmatrix}
		\dot{r}\\
		r\dot{\theta}\cos\phi\\
		r\dot{\phi}
	\end{bmatrix}
	=
	\begin{bmatrix}
		\cos\theta\cos\phi & \sin\theta\cos\phi & \sin\phi\\
		-\sin\theta & \cos\theta & 0\\
		-\cos\theta\sin\phi & -\sin\theta\sin\phi & \cos\phi
	\end{bmatrix}
	\begin{bmatrix}
		\dot{x}\\
		\dot{y}\\
		\dot{z}
	\end{bmatrix}
\end{gather}

Consider also spherical unit vectors $\mathbf{e}_r, \mathbf{e}_\theta, \mathbf{e}_\phi$ associated with a stationary feature $\mathbf{F}$ in the 3-D scene. Note that there are conversions from spherical unit vectors ($\mathbf{e}_r, \mathbf{e}_\theta, \mathbf{e}_\phi$) to/from rectilinear unit vectors ($\mathbf{i}, \mathbf{j}, \mathbf{k}$):

\begin{flushleft}
\end{flushleft}
\begin{gather}\label{unitVectorsRectToSpherical}
	\begin{bmatrix}
		\mathbf{e}_r\\
		\mathbf{e}_\theta\\
		\mathbf{e}_\phi
	\end{bmatrix}
	=
	\begin{bmatrix}
		\cos\theta\cos\phi & \sin\theta\cos\phi & \sin\phi\\
		-\sin\theta & \cos\theta & 0\\
		-\cos\theta\sin\phi & -\sin\theta\sin\phi & \cos\phi
	\end{bmatrix}
	\begin{bmatrix}
		\mathbf{i}\\
		\mathbf{j}\\
		\mathbf{k}
	\end{bmatrix}
\end{gather}
\begin{flushleft}
\end{flushleft}
\begin{gather}\label{unitVectorsSphericalToRect}
	\begin{bmatrix}
		\mathbf{i}\\
		\mathbf{j}\\
		\mathbf{k}
	\end{bmatrix}
	=
	\begin{bmatrix}
		\cos\theta\cos\phi & -\sin\theta & -\cos\theta\sin\phi\\
		\sin\theta\cos\phi & \cos\theta & -\sin\theta\sin\phi\\
		\sin\phi & 0 & \cos\phi	
	\end{bmatrix}
	\begin{bmatrix}
		\mathbf{e}_r\\
		\mathbf{e}_\theta\\
		\mathbf{e}_\phi
	\end{bmatrix}
\end{gather}
\\
We can now decompose $\mathbf{V}$ by its components using the convenient unit vectors $\mathbf{e}_r,\mathbf{e}_\theta,\mathbf{e}_\phi$ :

% VELOCITY EQUATION WITH UNIT VECTORS
\begin{align}
	\mathbf{V} &= \dot{r}\mathbf{e}_r + r\dot{\theta}\cos(\phi)\mathbf{e}_\theta + r\dot{\phi}\mathbf{e}_\phi \label{E:VelocityFieldSpherical}
\end{align}

Note that equations \eqref{E:VelocityField} and \eqref{E:VelocityFieldSpherical} refer to the same \textit{relative velocity field} $\mathbf{V}$.

% DERIVATION OF LOOMING FROM VELOCITY FIELD IN SPHERICAL COORDINATES
\paragraph{Looming from normalized velocity field}
We can get another expression for looming ($L$) by normalizing the \textit{relative velocity field} $\mathbf{V}$ by $r$. Dividing equation \eqref{E:VelocityField} by $r$ and expanding $\mathbf{t}$ and $\Omega$ using spherical unit vectors yield:

% EQUATIONS V/R EXPANDED LEFT
\begin{align}
	\frac{\mathbf{V}}{r} &= 
	\left(\frac{-\mathbf{t}}{r}\right) + \left(\frac{{-{\Omega} \times \mathbf{r}}}{r}\right)\notag\\
	&=\frac{-(t_r\mathbf{e}_r + t_\theta\mathbf{e}_\theta + t_\phi\mathbf{e}_\phi )}{r} \notag\\ 
	&- (\Omega_r\mathbf{e}_r + \Omega_\theta\mathbf{e}_\theta + \Omega_\phi\mathbf{e}_\phi ) \times \mathbf{e}_r \notag\\
	&=\left(\frac{-t_r}{r}\right)\mathbf{e}_r + \left(\frac{-t_\theta}{r}\right)\mathbf{e}_\theta + 
	\left(\frac{-t_\phi}{r}\right)\mathbf{e}_\phi \notag\\
	&+ 
	(-\Omega_\phi) \mathbf{e}_\theta +
	\Omega_\theta \mathbf{e}_\phi \label{E:VelocityFieldLeft}
\end{align}

where:
\begin{align}
	t_r &= \mathbf{t}\cdot \mathbf{e}_r ,&&\Omega_r = {\Omega}\cdot \mathbf{e}_r\notag\\
	t_\theta &= \mathbf{t}\cdot \mathbf{e}_\theta ,&&\Omega_\theta = {\Omega}\cdot \mathbf{e}_\theta\notag\\
	t_\phi &= \mathbf{t}\cdot \mathbf{e}_\phi ,&&\Omega_\phi = {\Omega}\cdot \mathbf{e}_\phi\notag
\end{align}

In addition, by dividing equation \eqref{E:VelocityFieldSpherical} by $r$ we obtain:

% EQUATION V/R EXPANDED RIGHT
\begin{equation}
	\frac{\mathbf{V}}{r}
	= \left(\frac{\dot{r}}{r}\right)\mathbf{e}_r + 
	\dot{\theta}\cos(\phi)\mathbf{e}_\theta + 
	\dot{\phi}\mathbf{e}_\phi\label{E:VelocityFieldSphericalRight}
\end{equation}

By equating \eqref{E:VelocityFieldLeft} and \eqref{E:VelocityFieldSphericalRight} and by isolating only the resultant $\mathbf{e}_r$ components on both sides of these equations we get:

% EQUATION EQUATING BOTH SIDES
\begin{equation}
	\left(\frac{-t_r}{r}\right)\mathbf{e}_r
	= \left(\frac{\dot{r}}{r}\right)\mathbf{e}_r \label{E:looming_er_components}
\end{equation}

or 

\begin{equation}
	\left(\frac{\mathbf{t} \cdot \mathbf{e}_r}{r}\right)
	= -\left(\frac{\dot{r}}{r}\right) \label{E:looming_er_components2}
\end{equation}

Therefore, by substituting $L$ from \eqref{E:Looming} into \eqref{E:looming_er_components2} we obtain another expression for looming in spherical coordinates:

% EQUATION LOOMING FROM VELOCITY
\begin{equation}\label{E:LoomingAndVelocity}
	L = \frac{\mathbf{t} \cdot \mathbf{e}_r}{r} 
\end{equation}

Notice that by knowing the instantaneous translation vector of the vehicle $\mathbf{t}$ and range $r$, we can compute the looming value ($L$) for any point along the directional unit vector $\mathbf{e}_r$.

% FIGURE EXPANDING LOOMING
\begin{figure*}
	\centering
	{\epsfig{file = 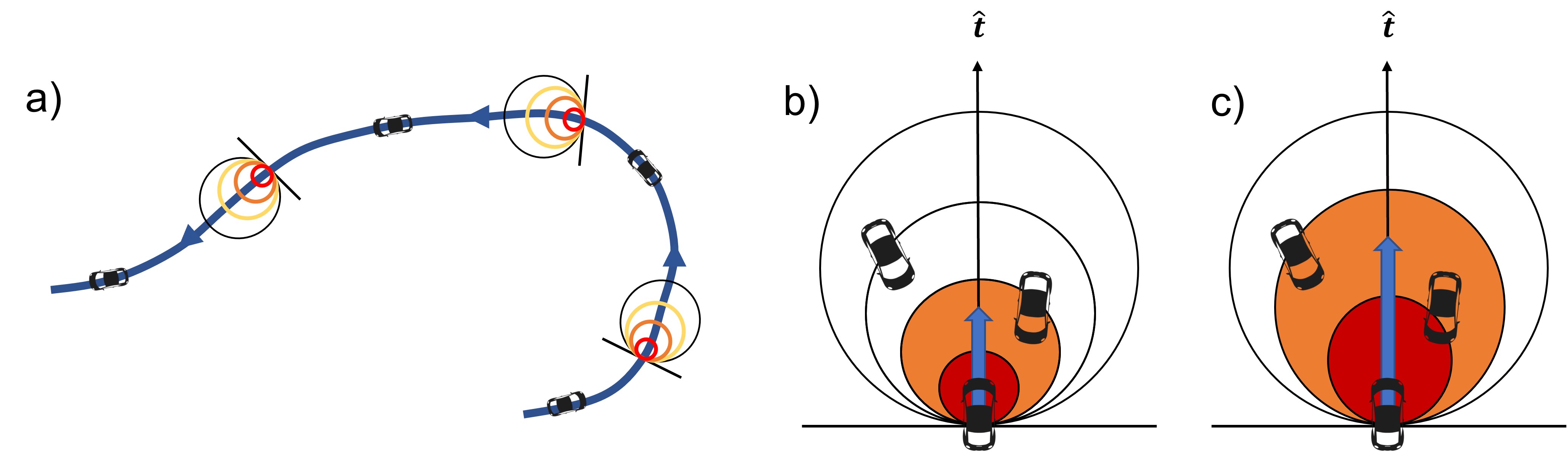, width = 15cm}}
	\caption{Equal looming circles. a) Vehicle trajectory, b) High and medium threat zones, c) Vehicle motion: higher speeds results in an expansion of the threat region. }
	\label{fig:ExpandingLooming}
\end{figure*}

\subsection{Looming from LiDAR}
We propose two ways to extract looming from LiDAR data.

\subsubsection{Using LiDAR data only}
Looming can be estimated using LiDAR point clouds.  Each full scan of the LiDAR results in a range image grid.  Using two consecutive scans we get two range image grids as obtained from the same 3D environment.  Theoretically, if the range value of each LiDAR pixel in each grid corresponds to the same 3D point in space, we obtain two range values from which the looming can be estimated, where the only error is due to the range values as obtained from the LiDAR measurements.  

However, this is practically not the case: the problem with this approach is that due to the vehicle motion the assumption may hold only when the changes from one image to another are infinitesimally small. To minimize the effect of incorrect looming calculations and improve the robustness of the approach, we use interpolation/decimation and discretization of the data in the grid allowing for range values that are closer to the real ones.  

We understand that this method adds error to the calculation of looming, but as can be seen in the results section it is possible to get some crude estimation of the looming values.  

Equation \eqref{E:looming_grids} shows the specific calculations.  

\begin{equation}
	L_{ij} = - \frac{ \left(\frac{r(\theta_i, \phi_i)_{j+1} - r(\theta_i, \phi_i)_j}{\Delta t}\right)}{r(\theta_i, \phi_i)_{j+1}}  \label{E:looming_grids}
\end{equation}

where $i = [1,2,..,n]$, $n$ is the number of points on the grid, $j+1, j$ correspond to the current and previous point cloud sample sets. The resulting $L_{ij}$ can be interpreted as a LiDAR-based looming image from where threat zones can be obtained for the purpose of collision avoidance actions.

\subsubsection{Using LiDAR + IMU}
Another method for obtaining looming is using LiDAR and IMU (Inertial Measurement Unit). Even though this is not the primary focus of this paper, its purpose is to demonstrate that LiDAR combined with IMU increase the robustness of computation of looming. However, this approach has its drawback because it gives incorrect values of looming for moving objects.
Note that it requires only the translation vector and not the rotation component of the LiDAR sensor. It uses ego-motion only partially, and there is no need for 3D reconstruction and image understanding.

We can compute looming ($L$) for every measured stationary point by using equation \eqref{E:LoomingAndVelocity}. The range $r$ is provided by the LiDAR sensor for every point on the point cloud. The instantaneous translation vector $\mathbf{t}$ of the vehicle can be obtained from the IMU or by other odometry methods. 

For special cases where vehicles have little side motion, like most automotive applications, the translation vector can be assumed to be equal to the vehicle forward motion component, and the magnitude (speed) can be obtained from the vehicle odometer or from the speed encoder of the wheel. 

\subsubsection{Threat maps}
From either method, threat regions can be obtained from looming values by assigning specific thresholds. For example, \textit{High}, \textit{Medium}, or \textit{Low} threat zones. These threat maps were obtained directly from measurements and are sufficient for identifying location threats. Object identification or 3D reconstruction is not necessary.  

Figure \ref{fig:ExpandingLooming}.a shows the looming sphere as a function of time for a given constant speed. Figures \ref{fig:ExpandingLooming}.b and \ref{fig:ExpandingLooming}.c show equal looming spheres using two threat zones. An increase in vehicle speed, as can be seen in Figure \ref{fig:ExpandingLooming}.c (compared to Figure \ref{fig:ExpandingLooming}.b) causes expansion of the time-based high threat zone and thereby includes new obstacles in these zones which were not deemed to be obstacles in Figure \ref{fig:ExpandingLooming}.b. 

This is in contrast to raw LiDAR data that provides range only and not threat values.

\section{RESULTS}
We present results of the methods for obtaining looming and threat zones using synthetic and real data.
% FIGURE LOOMING WITH SYNTETHIC DATA
\begin{figure}
	\centering
	{\epsfig{file = 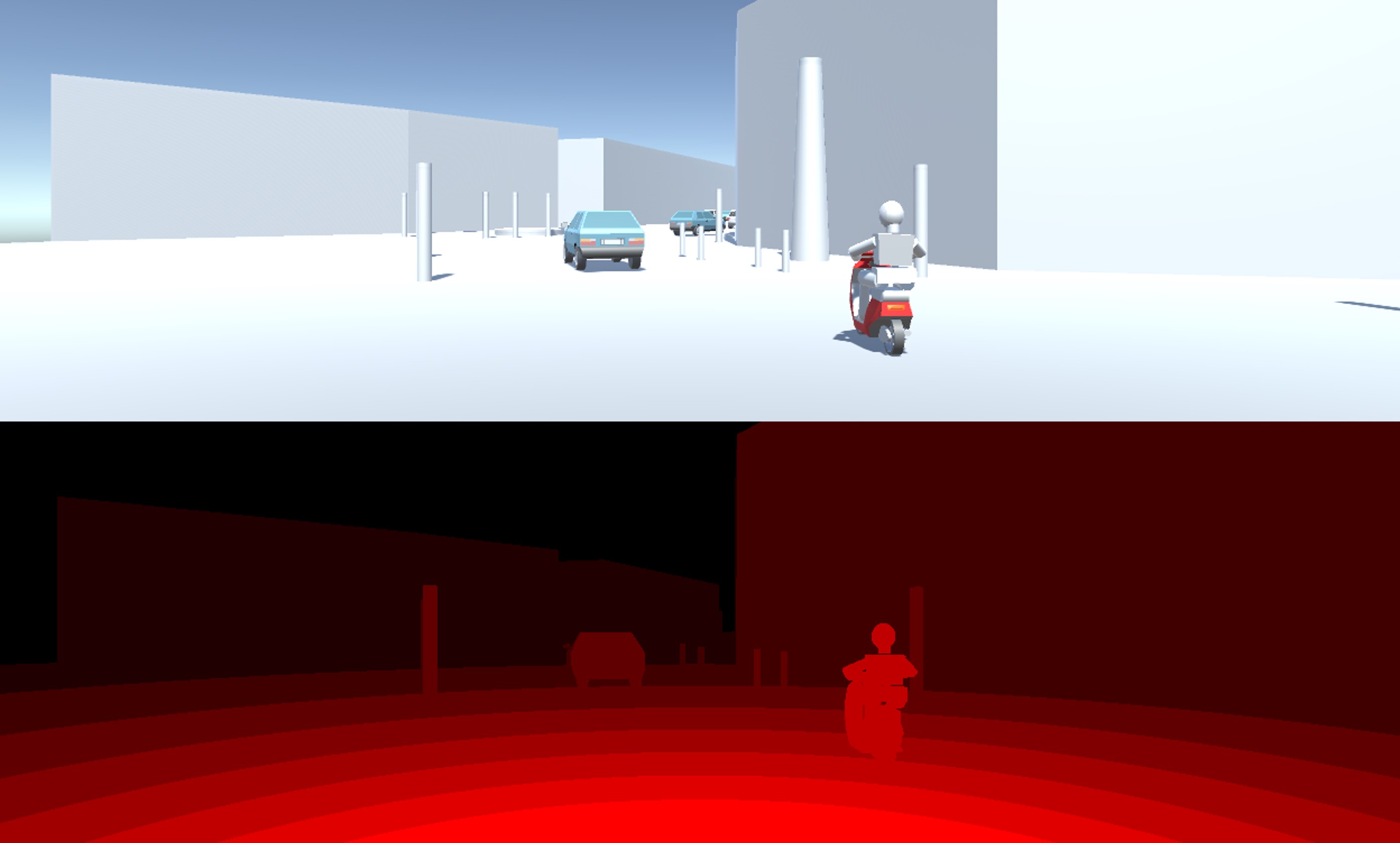, width = 7.8cm}}
	\caption{Looming from synthetic data: top image is the scene; bottom image is the looming map (Brighter red represents higher value of looming).}
	\label{fig:syntetic}
\end{figure}

% FIGURE RAW LIDAR DATA FROM KITTI
\begin{figure*}
	\centering
	{\epsfig{file = 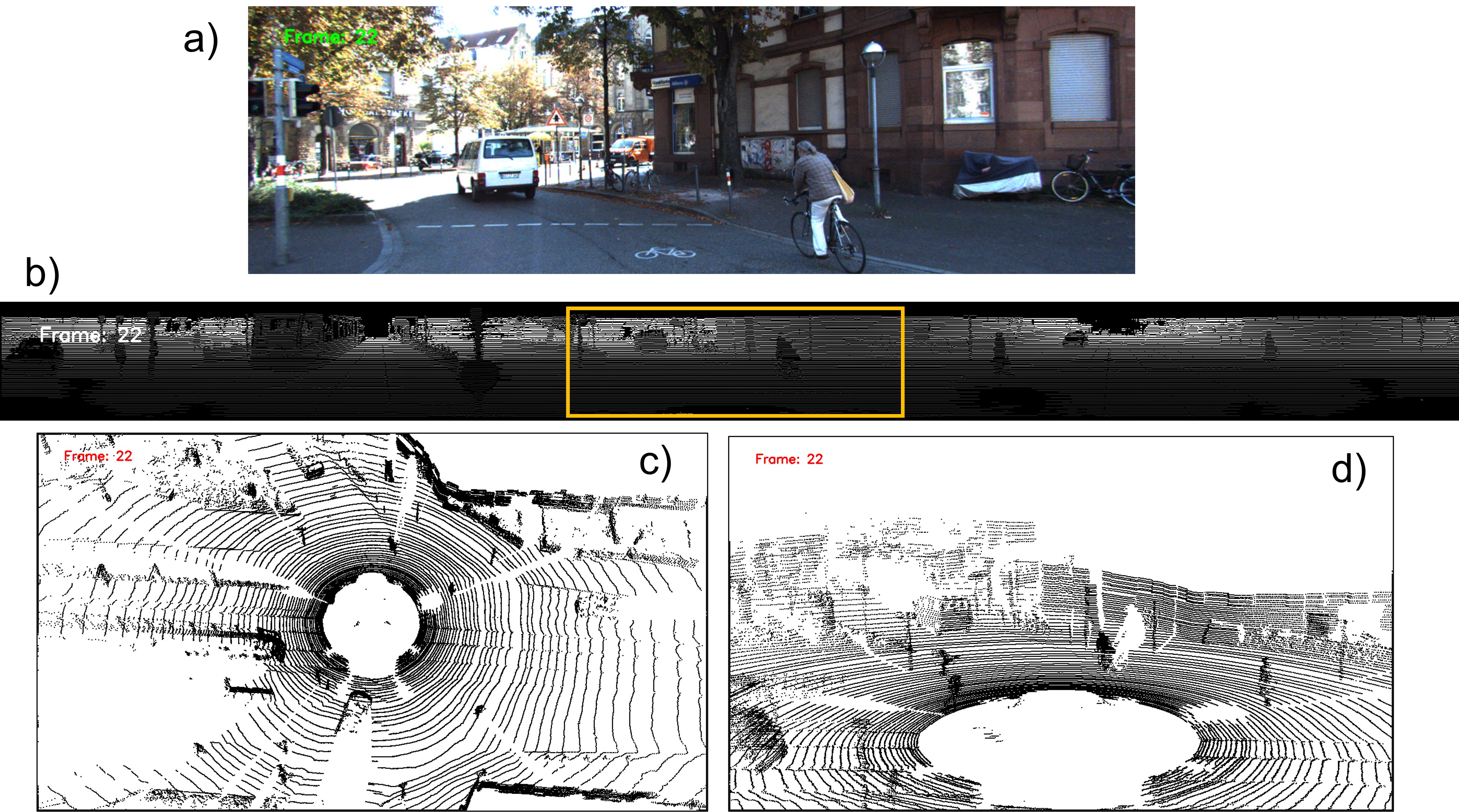, width = 15cm}}
	\caption{LiDAR data from the KITTI dataset: a) Original image from a color camera mounted on the vehicle, b) 360 degrees LiDAR data, the yellow rectangle corresponds to the color camera field of view, c) Top view of point cloud in 3D, d) Bird's eye 3D view of point cloud. }
	\label{fig:lidarData}
\end{figure*}

% FIGURE LIDAR LOOMING IN 3D
\begin{figure*}
	\centering
	\epsfig{file = 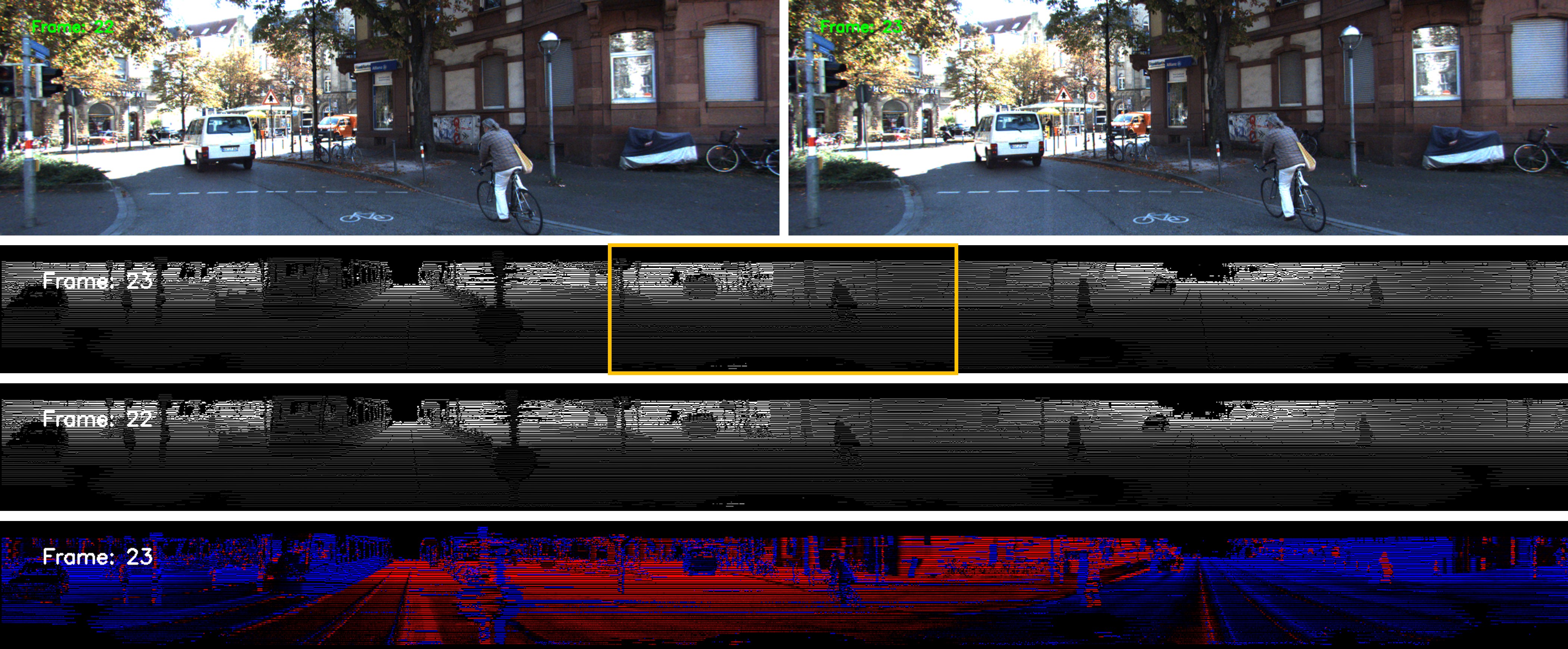, width = 15cm}
	\caption{Estimated looming from LiDAR only. (Refer to text for details).}
	\label{fig:LoomingFromGrids}
\end{figure*}

% FIGURE LIDAR LOOMING IN 3D
\begin{figure*}
	\centering
	\epsfig{file = 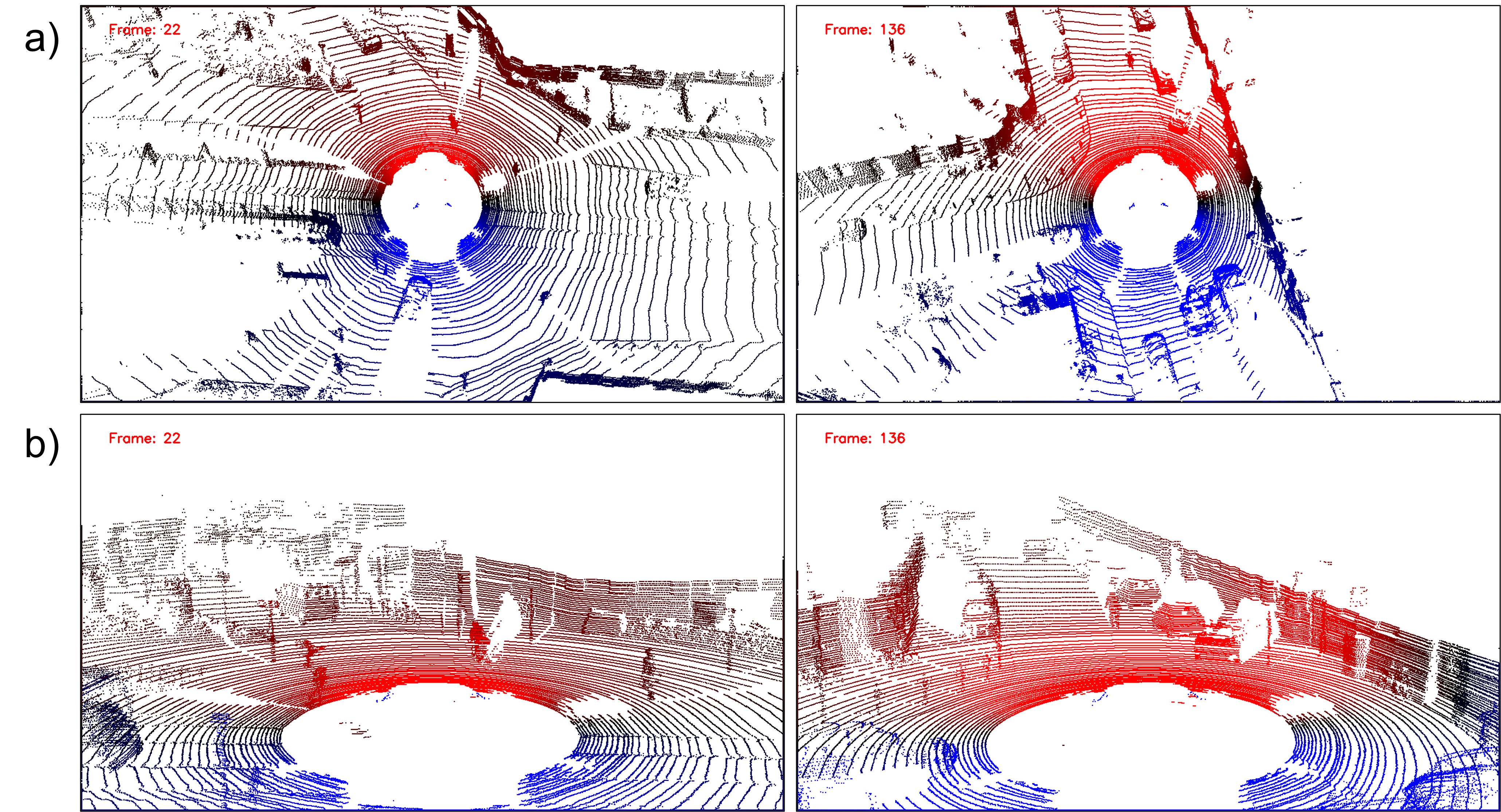, width = 15cm}
	\caption{Looming from LiDAR (red is potential threat): a) Top view, b) Bird's eye view. }
	\label{fig:Looming+IMU}
\end{figure*}

% FIGURE LIDAR LOOMING FRAMES 22, 136
\begin{figure*}
	\centering
	{\epsfig{file = 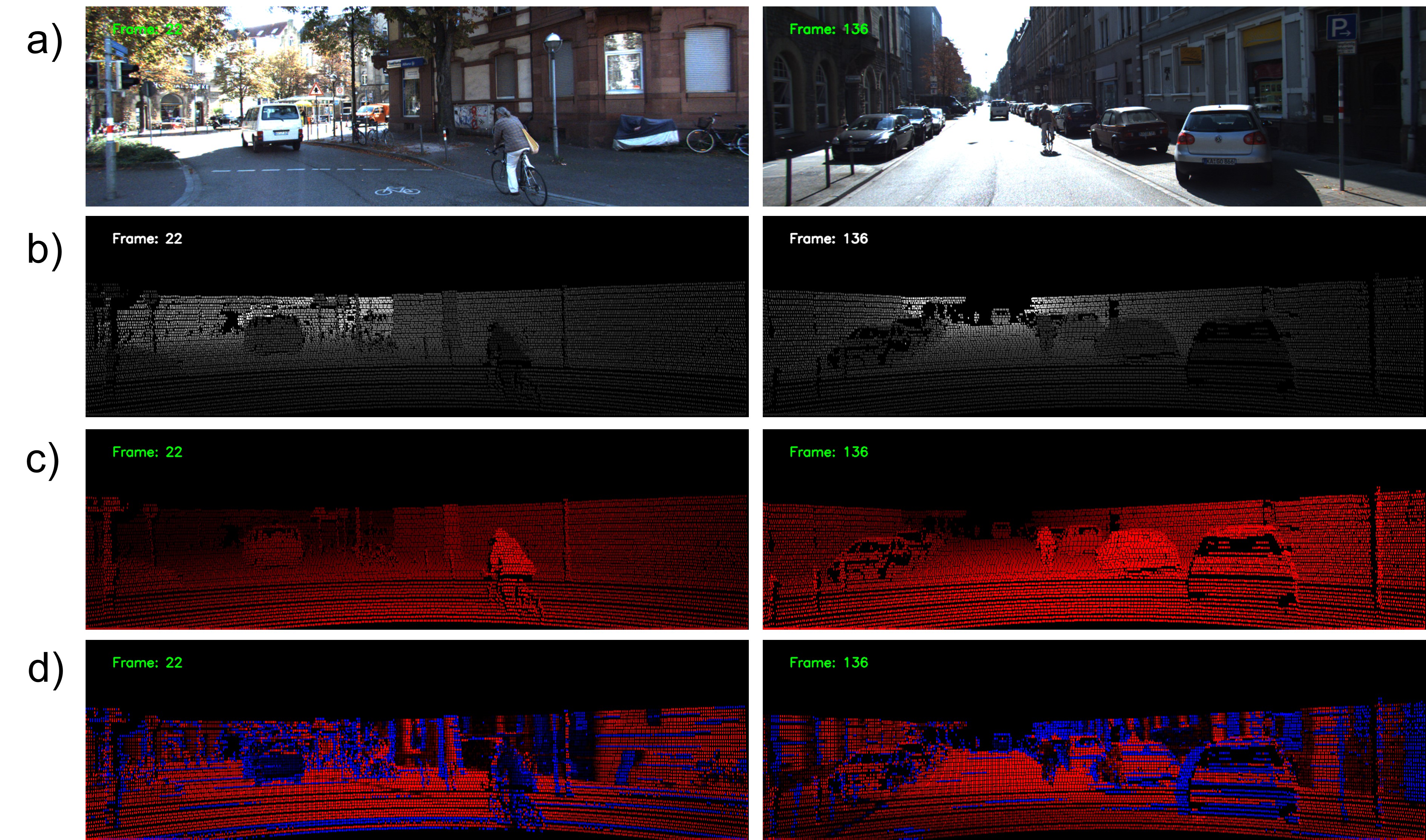, width = 15cm}}
	\caption{Looming from Lidar:  a) Original image, b) Depth from LiDAR, c) Looming from LiDAR + IMU, d) Estimated looming from LiDAR only.}
	\label{fig:LoomingEstimation}
\end{figure*}

% FIGURE THREAT ZONES
\begin{figure*}
	\centering
	{\epsfig{file = 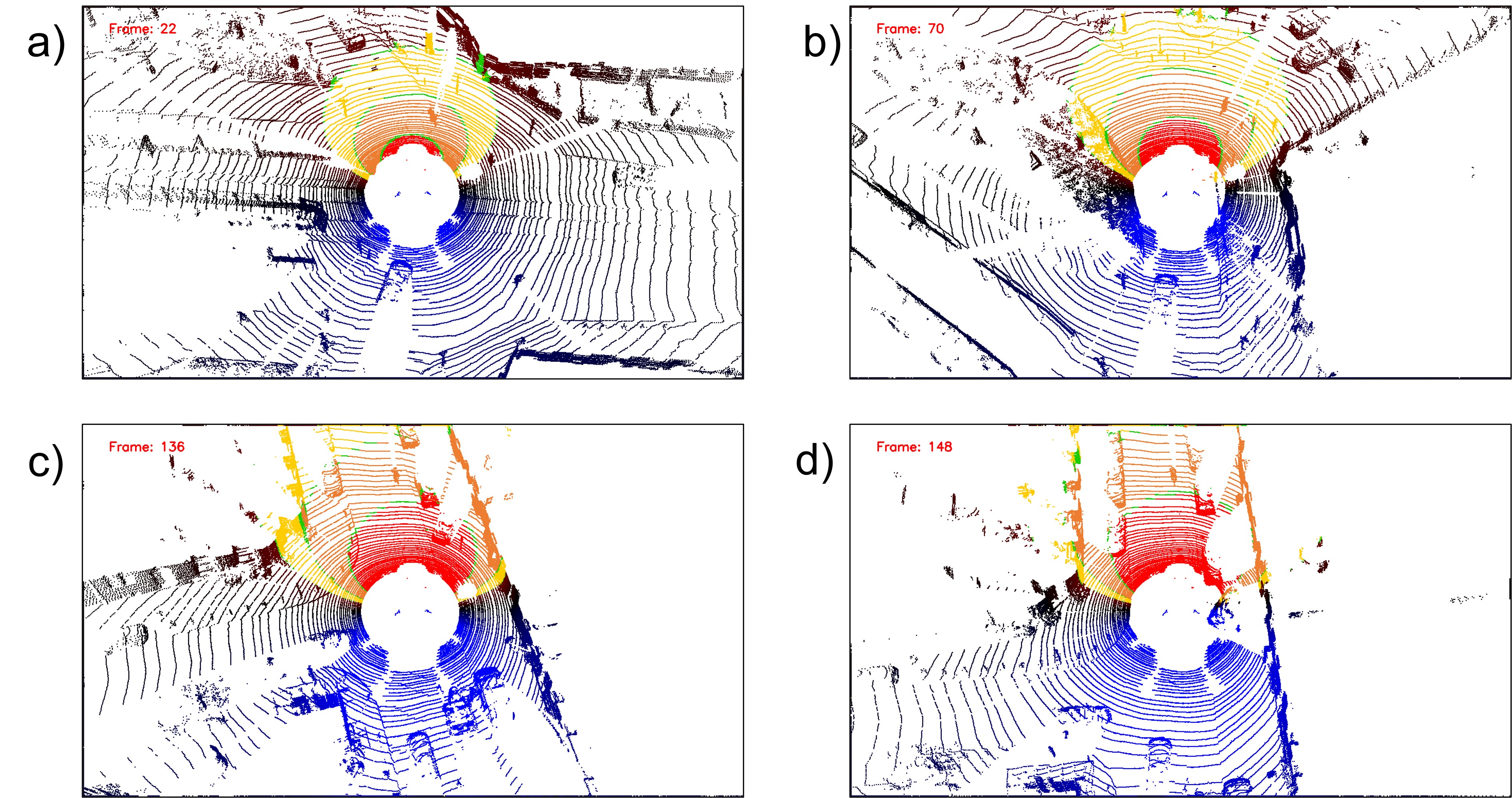, width = 15cm}}
	\caption{Threat zones expanding and contracting according to looming (refer to text for details) }
	\label{fig:lidarThreatZones}
\end{figure*}

\subsection{Ground truth looming with synthetic data }
A simulation of a moving vehicle in a stationary environment was performed using the Unity3D game engine. In this simulation the vehicle is translating forward at a constant speed between some stationary objects. The top part of Figure \ref{fig:syntetic} shows an actual image of the scene. The bottom part of Figure \ref{fig:syntetic} shows ground truth looming computed using equation \eqref{E:Looming} and shown from the observer point of view. A pixel shader filter was implemented to visualize the looming value as levels of red color. Brighter red color corresponds to higher value of looming.

\subsection{LiDAR data from KITTI dataset}
We processed real data using a particular city drive from the well-known KITTI dataset \cite{geiger2013vision}. The KITTI dataset includes raw data provided by a Velodyne 3D laser scanner (LiDAR) along with velocity vector from a GPS/IMU inertial navigation system. Color camera images were only used as reference as shown in Figure \ref{fig:lidarData}.a.
The LiDAR sensor specifications are:
Velodyne HDL-64E rotating 3D laser scanner, 10 Hz, 64 beams, 0.09-degree angular resolution, 2 cm distance, accuracy, provides around 1.3 million points/second, field of view: 360 degrees horizontal, 26.8 degrees vertical, range:120 m.

Shown in Figure \ref{fig:lidarData} are the original images from one of the vehicle color cameras and gray-scale representations of the raw LiDAR data. A yellow rectangle indicates where the LiDAR data matches the field of view of the color camera. Also shown are top and bird's eye views of the data for a specific time instant of the video.

\subsection{Estimation of looming using LiDAR}
In Figure \ref{fig:LoomingFromGrids}, original images from two consecutive frames (frames 22 and 23) are shown along LiDAR views (360 degrees) for two time instants. The resultant looming estimation using equation \eqref{E:looming_grids} is also shown at the bottom of Figure \ref{fig:LoomingFromGrids}. Notice positive values of looming are shown in red and corresponds mainly to points along the forward direction of motion (around the center of the image).  

Due to the effect of occlusions of points, there are sudden changes in range between frames, causing an "edge-effect" around objects with incorrect looming values, shown in intense red or blue colors.  \textit{However an important advantage is that the method incorporates values of looming as obtained from moving objects.} For example, the bicycle in the middle of the image is correctly portrayed with dark colors, meaning low values of looming as expected for this kind of motion since the bicycle velocity almost matches the velocity of the vehicle, and the rate of change in range is close to zero.

\subsection{Looming from LiDAR and IMU sensors}
\label{sec:realLiDAR+IMU}
We use equation \eqref{E:LoomingAndVelocity} to compute looming for each point. From the IMU sensor we obtained the translation velocity vector $\mathbf{t}$ and from the LiDAR sensor we obtained the range $r$.
Figure \ref{fig:Looming+IMU} shows top and bird's eye views for two time instants. Color intensity was assigned to each point representing the obtained looming value: red for positive values, and blue for negative values. 

We can see that the looming is very similar to the expected shown in simulation. \textit{This method adequately registers looming for stationary objects; however, it is imprecise for moving objects since the relative speed is not considered}.

In Figure \ref{fig:LoomingEstimation} a comparison between both methods is shown.

\subsection{Threat zones for collision free navigation}
In Figure \ref{fig:lidarThreatZones} several threat zones are computed using looming from section \ref{sec:realLiDAR+IMU}: \textit{red zone} - high threat, \textit{orange zone} - medium threat and \textit{yellow zone }- low threat. 
Notice how the threat zones change sizes between frames, this is due to the change in speed of the vehicle.

\section{\uppercase{Conclusions}}
\label{sec:conclusion}

Researchers and practitioners use LiDAR technology for autonomous navigation tasks. Sensory data from LiDAR systems provide point clouds that can be used for 3D reconstruction. Combined with information about ego-motion it can lead to scene understanding (using mainly machine learning and AI techniques) followed by path planning to achieve obstacle avoidance.

In this paper we demonstrate how to compute looming \textit{directly} from raw LiDAR data without 3D reconstruction. There is no need for scene understanding such as identifying cars, bikes, or pedestrians. Two approaches are shown, one uses LiDAR data only, and the other uses LiDAR data combined with IMU.

The approach shows that looming, which is measured in time units, provides information about imminent threat that can potentially be used for navigation tasks such as obstacle avoidance.  

When using only LiDAR for obtaining looming (i.e., using multiple LiDAR range images at multiple time instants) relative motion between the vehicle and the environment can yield good approximation of looming \textit {even in the presence of moving objects}. However, when using instantaneous LiDAR information and ego-motion information (using, for example, GPS data of the moving LiDAR sensor) the looming values that are obtained are better when moving relative to a stationary environment but are \textit {incorrect when moving objects are present}. The reason is that in the latter case motions of moving objects are not accounted for.  It appears that combining the two methods can further improve the results. 

The paper shares highly encouraging initial results and should be considered as “work in progress” as more looming from LiDAR-related methods are being explored.

\section*{\uppercase{Acknowledgements}}
The authors thank \textcolor{black}{Dr. Sridhar Kundur} for many fruitful discussion and suggestions as well as very detailed comments and clarifications that led to meaningful improvements of this manuscript.

\bibliographystyle{apalike}
{\small
	\bibliography{LidarLooming}}

\begin{thebibliography}{}

\bibitem[Ache et~al., 2019]{ache2019neural}
Ache, J.~M., Polsky, J., Alghailani, S., Parekh, R., Breads, P., Peek, M.~Y.,
  Bock, D.~D., von Reyn, C.~R., and Card, G.~M. (2019).
\newblock Neural basis for looming size and velocity encoding in the drosophila
  giant fiber escape pathway.
\newblock {\em Current Biology}, 29(6):1073--1081.

\bibitem[Albus and Hong, 1990]{albus1990motion}
Albus, J.~S. and Hong, T.~H. (1990).
\newblock Motion, depth, and image flow.
\newblock In {\em Proceedings., IEEE International Conference on Robotics and
  Automation}, pages 1161--1170. IEEE.

\bibitem[Aloimonos, 1992]{aloimonos1992visual}
Aloimonos, Y. (1992).
\newblock Is visual reconstruction necessary? obstacle avoidance without
  passive ranging.
\newblock {\em Journal of Robotic Systems}, 9(6):843--858.

\bibitem[Evans et~al., 2018]{evans2018synaptic}
Evans, D.~A., Stempel, A.~V., Vale, R., Ruehle, S., Lefler, Y., and Branco, T.
  (2018).
\newblock A synaptic threshold mechanism for computing escape decisions.
\newblock {\em Nature}, 558(7711):590--594.

\bibitem[Geiger et~al., 2013]{geiger2013vision}
Geiger, A., Lenz, P., Stiller, C., and Urtasun, R. (2013).
\newblock Vision meets robotics: The kitti dataset.
\newblock {\em The International Journal of Robotics Research},
  32(11):1231--1237.

\bibitem[Grigorescu et~al., 2020]{grigorescu2020survey}
Grigorescu, S., Trasnea, B., Cocias, T., and Macesanu, G. (2020).
\newblock A survey of deep learning techniques for autonomous driving.
\newblock {\em Journal of Field Robotics}, 37(3):362--386.

\bibitem[Kundur and Raviv, 1999]{kundur1999novel}
Kundur, S.~R. and Raviv, D. (1999).
\newblock Novel active vision-based visual threat cue for autonomous navigation
  tasks.
\newblock {\em Computer Vision and Image Understanding}, 73(2):169--182.

\bibitem[Mujumdar and Padhi, 2011]{mujumdar2011evolving}
Mujumdar, A. and Padhi, R. (2011).
\newblock Evolving philosophies on autonomous obstacle/collision avoidance of
  unmanned aerial vehicles.
\newblock {\em Journal of Aerospace Computing, Information, and Communication},
  8(2):17--41.

\bibitem[Raviv, 1992]{raviv1992quantitative}
Raviv, D. (1992).
\newblock {\em A quantitative approach to looming}.
\newblock US Department of Commerce, National Institute of Standards and
  Technology.

\bibitem[Raviv and Joarder, 2000]{Raviv2000TheVL}
Raviv, D. and Joarder, K. (2000).
\newblock The visual looming navigation cue: A unified approach.
\newblock {\em Comput. Vis. Image Underst.}, 79:331--363.

\bibitem[Ridwan, 2018]{ridwan2018looming}
Ridwan, I. (2018).
\newblock {\em Looming object detection with event-based cameras}.
\newblock University of Lethbridge (Canada).

\bibitem[Roriz et~al., 2021]{roriz2021automotive}
Roriz, R., Cabral, J., and Gomes, T. (2021).
\newblock Automotive lidar technology: A survey.
\newblock {\em IEEE Transactions on Intelligent Transportation Systems}.

\bibitem[Sharma et~al., 2021]{sharma2021recent}
Sharma, O., Sahoo, N.~C., and Puhan, N. (2021).
\newblock Recent advances in motion and behavior planning techniques for
  software architecture of autonomous vehicles: A state-of-the-art survey.
\newblock {\em Engineering applications of artificial intelligence},
  101:104211.

\bibitem[Wang et~al., 2018]{wang2018robust}
Wang, M., Voos, H., and Su, D. (2018).
\newblock Robust online obstacle detection and tracking for collision-free
  navigation of multirotor uavs in complex environments.
\newblock In {\em 2018 15th International Conference on Control, Automation,
  Robotics and Vision (ICARCV)}, pages 1228--1234. IEEE.

\bibitem[Yasin et~al., 2020]{yasin2020unmanned}
Yasin, J.~N., Mohamed, S.~A., Haghbayan, M.-H., Heikkonen, J., Tenhunen, H.,
  and Plosila, J. (2020).
\newblock Unmanned aerial vehicles (uavs): Collision avoidance systems and
  approaches.
\newblock {\em IEEE access}, 8:105139--105155.

\bibitem[Yilmaz and Meister, 2013]{yilmaz2013rapid}
Yilmaz, M. and Meister, M. (2013).
\newblock Rapid innate defensive responses of mice to looming visual stimuli.
\newblock {\em Current Biology}, 23(20):2011--2015.

\bibitem[Zhang and Singh, 2014]{zhang2014loam}
Zhang, J. and Singh, S. (2014).
\newblock Loam: Lidar odometry and mapping in real-time.
\newblock In {\em Robotics: Science and Systems}, volume~2, pages 1--9.
  Berkeley, CA.

\end{thebibliography}

\end{document}